%% file: main.tex
\documentclass[sigconf]{acmart}

\usepackage{multirow}
\usepackage[ruled,vlined]{algorithm2e}
\usepackage{algorithmic}
\usepackage{amsmath}
\usepackage{xspace}

\AtBeginDocument{%
  \providecommand\BibTeX{{%
    \normalfont B\kern-0.5em{\scshape i\kern-0.25em b}\kern-0.8em\TeX}}}

\setcopyright{rightsretained} 

\newcommand{\para}[1]{\paragraph{#1}}
\newcommand{\RR}{${\rm I\!R}$} 
\newcommand{\Ng}[1]{\mathcal{N}(#1)}
\newcommand{\Structack}{Structack\xspace}

\DeclareRobustCommand{\ex}[1]{#1}

\begin{document}
\fancyhead{}

\copyrightyear{2021}
\acmYear{2021}
\acmConference[HT '21]{Proceedings of the 32nd ACM Conference on Hypertext and Social Media}{August 30-September 2, 2021}{Virtual Event, Ireland}
\acmBooktitle{Proceedings of the 32nd ACM Conference on Hypertext and Social Media (HT '21), August 30-September 2, 2021, Virtual Event, Ireland}\acmDOI{10.1145/3465336.3475110}
\acmISBN{978-1-4503-8551-0/21/08}

\title{Structack: Structure-based Adversarial Attacks \\ on Graph Neural Networks}

\author{Hussain Hussain}
\email{hussain@tugraz.at}
\affiliation{%
  \institution{Graz University of Technology}
  \country{Austria}
}
\affiliation{%
  \institution{Know Center GmbH}
  \country{Austria}
}
\author{Tomislav Duricic}
\email{tduricic@tugraz.at}
\affiliation{%
  \institution{Graz University of Technology}
  \country{Austria}
}
\affiliation{%
  \institution{Know Center GmbH}
  \country{Austria}
}
\author{Elisabeth Lex}
\email{elisabeth.lex@tugraz.at}
\affiliation{%
  \institution{Graz University of Technology}
  \country{Austria}
}
\author{Denis Helic}
\email{dhelic@tugraz.at}
\affiliation{%
  \institution{Graz University of Technology}
  \country{Austria}
}
\author{Markus Strohmaier}
\email{markus.strohmaier@cssh.rwth-aachen.de}
\affiliation{%
  \institution{RWTH Aachen}
  \country{Germany}
}
\affiliation{%
  \institution{GESIS - Leibniz Institute for the Social Sciences}
  \country{Germany}
}
\author{Roman Kern}
\email{rkern@tugraz.at}
\affiliation{%
  \institution{Graz University of Technology}
  \country{Austria}
}
\affiliation{%
  \institution{Know Center GmbH}
  \country{Austria}
}

\renewcommand{\shortauthors}{Hussain, et al.}


\begin{abstract}
Recent work has shown that graph neural networks (GNNs) are vulnerable to adversarial attacks on graph data.
Common attack approaches are typically \textit{informed}, i.e. they have access to information about node attributes such as labels and feature vectors.
In this work, we study adversarial attacks that are \textit{uninformed}, where an attacker only has access to the graph structure, but no information about node attributes.
Here the attacker aims to exploit structural knowledge and assumptions, which GNN models make about graph data.
In particular, literature has shown that structural node centrality and similarity have a strong influence on learning with GNNs.
Therefore, we study the impact of centrality and similarity on adversarial attacks on GNNs.
We demonstrate that attackers can exploit this information to decrease the performance of GNNs by focusing on injecting links between nodes of low similarity and, surprisingly, low centrality.
We show that structure-based uninformed attacks can approach the performance of informed attacks, while being computationally more efficient.
With our paper, we present a new attack strategy on GNNs that we refer to as \Structack.
\Structack can successfully manipulate the performance of GNNs with very limited information while operating under tight computational constraints.
Our work contributes towards building more robust machine learning approaches on graphs.

\end{abstract}




\maketitle

\input{introduction}

\input{preliminaries}
\input{model}

\input{experiments}
\input{discussion}

\input{related-work}

\input{conclusion}
\input{acknowledgements}

\bibliographystyle{ACM-Reference-Format}
\bibliography{references}

\newpage
\appendix
\input{appendix}

\end{document}

%% file: introduction.tex
\section{Introduction}

\begin{figure*}[t]
\centering
\includegraphics[width=0.9\linewidth]{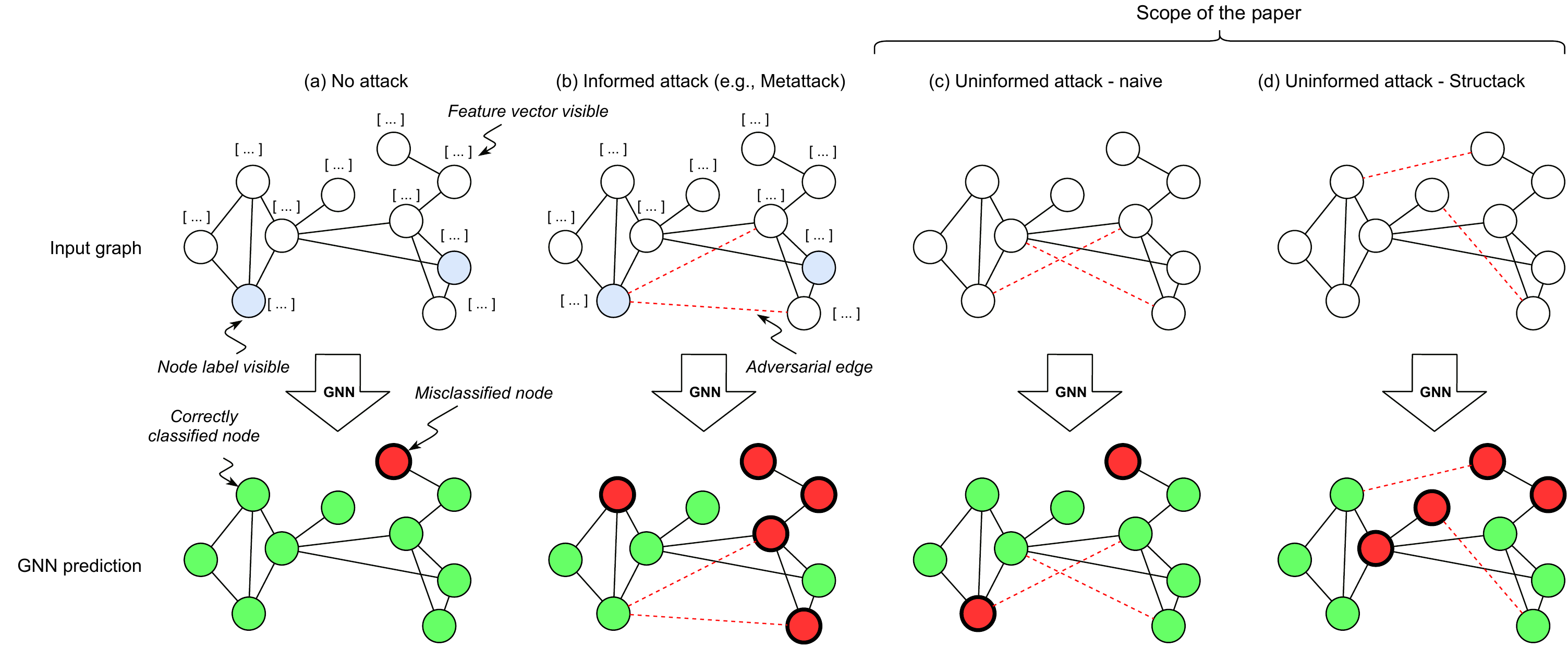}
\caption{Illustration of \Structack on GNN classification. In (a) we show a standard GNN classification with all node features and labels available to the algorithm. In (b) we depict an informed attack, which has access to the same information that the GNN classification task itself has. Based on this information, adversarial edges are added to attack GNN performance.
In (c), we show an uninformed attack strategy, i.e. an attack that has no access to information about node attributes (labels, features), but only to the structure of the graph. A naive strategy could add edges based on topological graph features, for example add edges between pairs of nodes with high centrality or high similarity.
In (d), we show a \Structack attack which also has no access to information about node attributes, but attacks more successfully by adding edges between nodes with low centrality and low similarity.
\Structack approaches the performance of informed attacks such as Metattack~\cite{zugner2019metattack} with less available information.
}
\label{fig:teaser}
\end{figure*}

 
Graph neural networks (GNNs) are state-of-the-art models for tasks on graphs such as node classification~\cite{gcn}, link prediction~\cite{zhang2018link} and graph classification~\cite{graphsage}.
Recent work has shown that GNNs are vulnerable to adversarial attacks, which can cause GNNs to fail by carefully manipulating node attributes~\cite{ma2020practical}, graph structure~\cite{ma2019rewatt,zugner2019metattack} or both~\cite{zugner2018nettack}.
For example, adversarial attacks on social networks can add links via fake accounts, or change the personal data of a controlled account.
Most existing attacks~\cite{zugner2018nettack,zugner2019metattack,xu2019topology,dai2018rls2v,ma2019rewatt,ma2020practical} assume that information about node attributes (e.g., demographics of users) are available to the attacker. 
In practice however, attackers have limited access to such attribute information.
We thus differentiate between two cases: the \textit{informed} case where both graph structure and node attributes are available to the attacker, and the \textit{uninformed} case where only information about the structure is available (see Figure~\ref{fig:teaser}). 


\para{Objectives.}
    In this work, we investigate \emph{uninformed} adversarial attacks that aim to reduce the overall accuracy of node classification with GNNs by manipulating the graph structure. Our aim is to study (i) potential strategies for uninformed attacks and (ii) how effective they are in practical settings. 



\para{Approach.} 
Insights in~\cite{zugner2019metattack,ma2020practical,jin2020prognn} have shown a considerable influence of node degree and shortest paths on GNN robustness.
However, these insights are not well investigated.
Therefore, we further inspect the effect of degree centrality and shortest path lengths on GNN adversarial attacks.
    First, we theoretically show that with standard degree normalization, low-degree neighbors surprisingly have more influence on a node's representation than higher-degree neighbors.
    Second, we discuss the results showing the dependency of GNNs on links within graph communities~\cite{Li2018Deeper,hussain2020impact}, which are ubiquitous in real-world graphs.
    Based on that, we argue that adversarial edges should link nodes with longer paths between them.
    Experimentally, we verify these insights on degrees and distance through simulating attacks on empirical datasets.
    We then introduce our uninformed \textbf{struc}ture-based adversarial at\textbf{tack} (\Structack), which generalizes these findings, and injects links between nodes of low structural centrality and similarity.
    Finally, we evaluate \Structack compared to state-of-the-art attacks in terms of (i) reducing GNN accuracy, (ii) computational efficiency, and (iii) the ability to remain undetected.

\para{Contribution and Impact.}
    We introduce \Structack\footnote{We provide the implementation \Structack and the experiments for reproducibility at \url{https://github.com/sqrhussain/structack}.}, a novel structure-based uninformed adversarial attack on GNNs.
    In experiments on empirical datasets, \Structack performs on a level that is comparable to more informed state-of-the-art attacks~\cite{xu2019topology,zugner2019metattack}, while using less information about the graph and significantly lower computational requirements.
    We give insights on the detection of attacks, such as \Structack, by analyzing their ability to be undetected.
    With our work, we introduce a new unstudied category of attacks that could be applied to real-world networks.
    Our findings highlight the vulnerability of GNNs to uninformed attacks that have no knowledge about node attributes or the attacked model.
    Hence, our work contributes toward \ex{building} more robust predictive and defensive models for graph data.

%% file: preliminaries.tex
\section{Background}
\label{sec:prelim}
\begin{figure*}
    \includegraphics[width=0.95\linewidth]{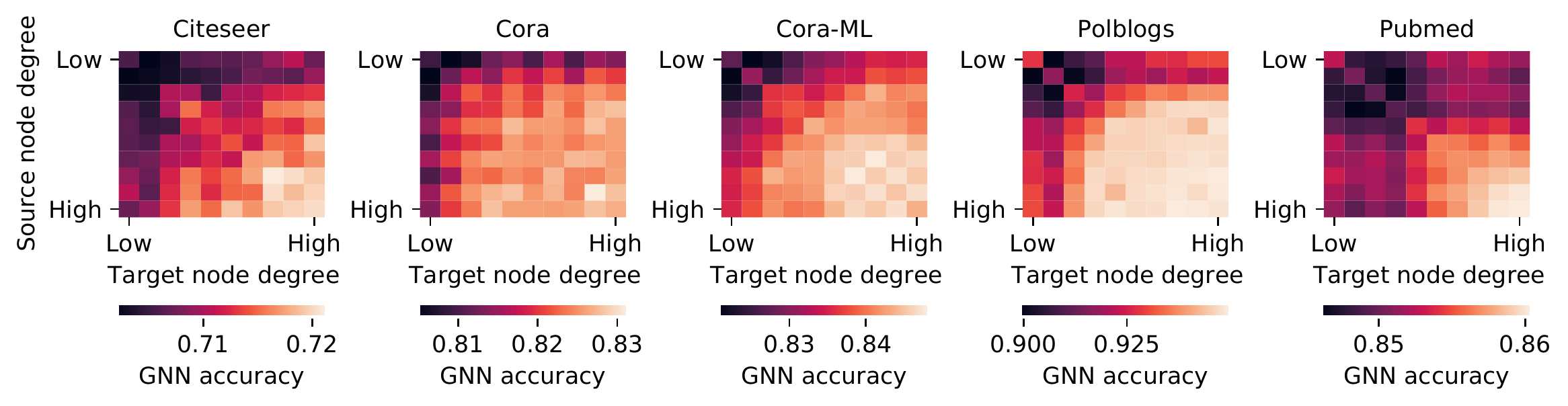}
    \includegraphics[width=.95\textwidth]{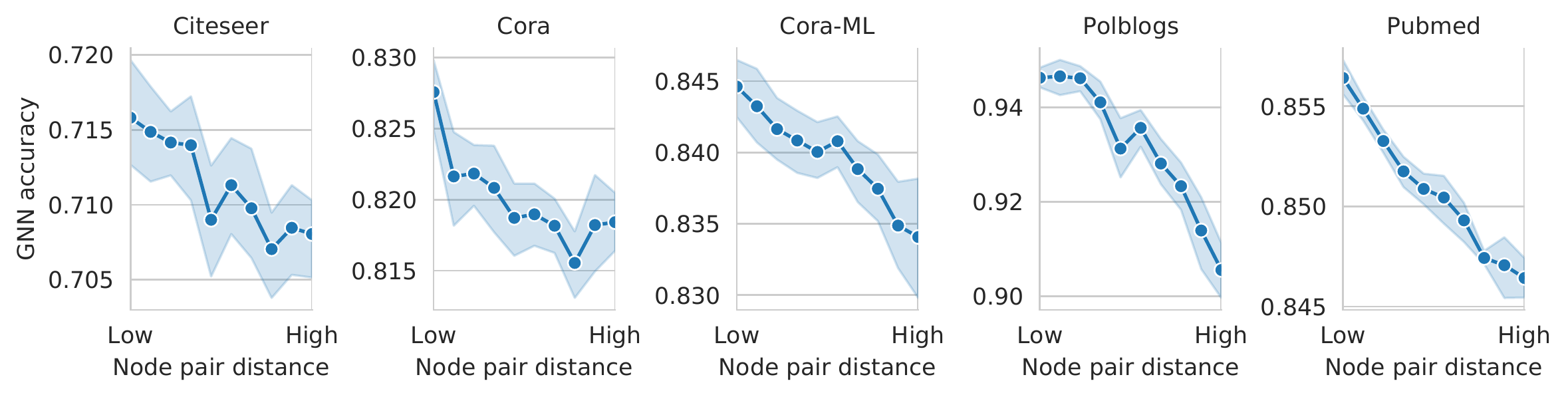}
    \caption{
    Impact of degree and distance on adversarial attacks on GNN classification.
    \textit{Top}: GNN accuracy when we link nodes of varying degrees to other nodes of varying degrees as well, i.e., low-to-low degrees (top-left corner) up to high-to-high (bottom-right corner).
    Linking nodes with lower node degrees appears to result in more effective attacks.
    \textit{Bottom}: GNN accuracy when adding edges between pairs of nodes with the lowest distance up to the highest distance.
    Linking nodes with higher distance (lower similarity) results in a more effective attack.
    Degrees and distances are grouped into 10-quantiles. The presented accuracy comes from training a GNN model (namely GCN\cite{gcn}) on the perturbed graphs of $5$ empirical datasets. 
    }
    \label{fig:injection-experiment}
\end{figure*}
\para{Preliminaries.}
Let $G=(A,X,Y)$ be an attributed undirected graph with an unweighted adjacency matrix $A \in \{0,1\}^{n \times n}$, a feature matrix $X \in \RR^{n \times f}$, and a label matrix $Y \in \{0,1\}^{n \times |L|}$, where $L$ is the set of labels. We refer to the set of nodes as $V=\{1,...,n\}$, and the set of edges as $E$, where $|E|=m$ and $(i,j)\in E$ \textit{iff} $A_{i,j}=1$.
Each node $u \in V$ has a feature vector $x_u \in \RR^f$, where $f$ is the feature vector dimension, and a label $y_u \in L$.
The feature vectors are encoded in $X$, where $u$'s feature vector $x_u^T$ is the row $u$ of matrix~$X$.
The labels of all the nodes are accordingly encoded in $Y$ as well, with one-hot encoding in each row.
We use the notation~$D$ to refer to the degree matrix, a diagonal matrix where $D_{i,i} = d_i$ is the degree of node~$i$.

\para{Graph neural networks.}
GNNs are multi-layer machine learning models that process graph-structured data.
They follow a message passing and aggregation scheme, where nodes aggregate the messages received from their neighbors and update their representation on this basis.
For a GNN with $K$ layers, \citet{gcn} describe the propagation rule in a simple form as follows 
\begin{equation}
    \label{eq:gcn}
    H'^{(k+1)} = \sigma (\Tilde{D}^{-\frac{1}{2}}\Tilde{A}\Tilde{D}^{-\frac{1}{2}} H'^{(k)} W^{(k)})
\end{equation}
for $k = 0,1,...,K-1$, where $\Tilde{A}=A+I$, $\Tilde{D}=D+I$, $H'^{(0)}=X$ are the given node features, $W^{(k)}$ is a trainable weight matrix, and $\sigma$ is an activation function, which is typically non-linear, e.g., ReLU.
This formula is usually written as $H'^{(k+1)} = \sigma (\Hat{A} H'^{(k)} W^{(k)})$  with $\Hat{A} = \Tilde{D}^{-\frac{1}{2}}\Tilde{A}\Tilde{D}^{-\frac{1}{2}}$.

\para{Node classification with GNNs.}
For node classification, we set the activation function of the last layer of the GNN to softmax
\begin{equation}
    \label{eq:softmax}
    Z = f(\Theta;A,X) = \text{softmax}(\Hat{A} H'^{(K-1)} W^{(K-1)}),
\end{equation}
where $\Theta=\{W^{(0)},...,W^{(K-1)}\}$ is the set of model parameters which we aim to optimize, and $Z_{u,c}$ represents the model confidence that a node $u$ belongs to label $c$.
Given the labels of a subset of nodes $V' \subseteq V$, the goal of node classification is to find the labels of the unlabeled nodes $V \setminus V'$.
To achieve this with GNNs, a common choice is to minimize the cross entropy error in $V'$
\begin{equation}
    \label{eq:loss}
    \mathcal{L}(Y,Z) = - \sum_{u \in V'}{\ln Z_{u,y_u}},
\end{equation}
where $Z_{u,y_u}$ represents the model confidence that node $u$ belongs to its ground-truth class $y_u$.

\para{Adversarial attacks on GNNs.}
GNNs are prone to global adversarial attacks on the graph structure (i.e., the adjacency matrix $A$)~\cite{jin2020survey}.
These attacks aim to reduce the overall (i.e., \textit{global}) node classification accuracy of a GNN model.
To that end, these attacks follow different strategies to perturb the graph structure by adding or removing up to a \textit{budget} of $k$ edges.

\section{Impact of the structure on attacks}
\label{sec:hypotheses}
This section presents examples of graph structural properties and analyzes their effect on adversarial attacks on graph structure. 
Findings from related work show that node degrees have an impact on graph controllability~\cite{liu2011controllability} and on the selection of targeted nodes in adversarial attacks on node attributes~\cite{ma2020practical}.
Besides, an analysis of Metattack (a state-of-the-art attack) in~\cite{zugner2019metattack} shows a slight tendency of the attack to link pairs of nodes with longer shortest paths (i.e., longer distances\footnote{We refer to shortest path lengths as distances for brevity.}).
However, these works do not particularly focus on the impact of node degrees and distances on adversarial attacks or the reasoning behind it. 
Therefore, in the following analysis, we study the impact of node degrees and distances on GNNs from a theoretical perspective, and consequently verify this impact empirically.

\subsection{Impact of degree and distance}
\label{sec:analysis}

\para{Node degree impact.}
\label{sec:node-degree}
We aim to theoretically assess the role of node degree on the propagation in GNNs (Equation~\ref{eq:gcn}).
For this study, we investigate the common degree normalization form as in Equation~\ref{eq:gcn}, i.e., normalization by the degree square root of two adjacent nodes. Using other less common forms of degree normalization or no degree normalization can be investigated in future work.
We first simplify the update rule given in Equation~\ref{eq:gcn} by ignoring the non-linearity in the intermediate layers, i.e., linearizing the equation (inspired by~\cite{zugner2018nettack} and~\cite{wu2019simplifying})
\begin{equation}
    \label{eq:linearized}
    H'^{(K)} := \text{softmax}(H^{(K)} W)= \text{softmax}(\Hat{A}^K X W),
\end{equation}
where weight matrices $W^{(k)}$ for $k \in \{0,1,..,K-1\}$ are absorbed by $W=W^{(0)} W^{(1)} ... W^{(K-1)} \in \RR^{f \times |L|}$.
We use $H^{(k)} = \Hat{A}^k X \in \RR^{n \times f}$ to represent node intermediate representations at layer $k$ in the linearized model.
Each row $u$ of matrix $H^{(k)}$, denoted as $(h_u^{(k)})^T \in \RR^{f}$, is the intermediate representation of node $u$ at layer $k$. 
As $H^{(k)} = \Hat{A} H^{(k-1)}$, we can write the representation in layer $k$ of node $u$ (i.e., $h_u^{(k)}$) in terms of the representations of its neighboring nodes $\Ng{u}$ in the previous layer $k-1$ as follows
\begin{equation}
    h_u^{(k)} = \sum_{v \in \Ng{u}}{\frac{1}{\sqrt{d_u d_{v}}} h_{v}^{(k-1)}}.\\
    \label{eq:recursion}
\end{equation}
To show the impact of the degree of a specific neighbor $w \in \Ng{u}$ on the node $u$, we compute the derivative of $u$'s final representation $h_u^{(K)}$ with respect to $w$'s initial representation $h_w^{(0)}$ (i.e., the input features for node $w$), that is, the Jacobian matrix $\mathbf{J}^{u,w} \in \RR^{f \times f}$ with $J_{i,j}^{u,w} = \partial h_{u,i}^{(K)}/\partial h_{w,j}^{(0)}$.
Equation~\ref{eq:recursion} shows that the $i$-th vector component of $h_u^{(k)} : k > 0$ (i.e., $h_{u,i}^{(k)}$) only depends on the vector component $h_{w,i}^{(k-1)}$ of the neighbor $w$, and not on any other component $h_{w,j}^{(k-1)}$ with $i \neq j$\footnote{This argument is possible due to linearizing the Equation~\ref{eq:gcn}.}.
By induction, we can show that, for $w \in \Ng{u}$, the $i$-th vector component of $h_{u}^{(k)}$ only depends on the $i$-th vector component of $h_{w}^{(0)}$.
This fact leads to the Jacobian matrix being diagonal. Therefore, it is sufficient to compute the partial of an arbitrary component~$i$

\begin{equation}
    J_{i,i}^{u,w}  = \frac{\partial h_{u,i}^{(K)}}{\partial h_{w,i}^{(0)}} = \frac{\partial (\sum_{v \in \Ng{u}}{\frac{1}{\sqrt{d_u d_{v}}} h_{v}^{(k-1)}})}{\partial h_{w,i}^{(0)}}. \\
\end{equation}
By applying the chain rule, we get
\begin{equation}
    \label{eq:generalization}
    \begin{split}
    J_{i,i}^{u,w} & = \sum_{v_1 \in \Ng{u}}{\frac{1}{\sqrt{d_u d_{v_1}}} \frac{\partial h_{v_1,i}^{(K-1)}}{\partial h_{w,i}^{(0)}}}\\
    \end{split}
\end{equation}
By repeatedly applying the chain rule $K$ times, we end up at the partial of a node's initial representation $h_{v_K,i}^{(0)}$ in terms of $w$'s initial representation $h_{w,i}^{(0)}$, that is
\begin{equation}
    \label{eq:base}
    \frac{\partial h_{v_K,i}^{(0)}}{\partial h_{w,i}^{(0)}} =
    \begin{cases}
    1 : v_K = w\\
    0 : v_K \neq w.
    \end{cases}
\end{equation}
When we propagate this back to Equation~\ref{eq:generalization}, we arrive at
\begin{equation}
    \label{eq:generalization-ext}
    \begin{split}
    J_{i,i}^{u,w}  & = \sum_{v_1 \in \Ng{u}}{\frac{1}{\sqrt{d_u d_{v_1}}} (... (\sum_{v_K \in \{w\}}{\frac{1}{\sqrt{d_{v_{K-1}} d_{v_K}}}}) ...)}\\
    \end{split}
\end{equation}
We can rewrite Equation~\ref{eq:generalization-ext} for each (not necessarily simple) path of length $K$ between $u$ and $w$, that is, with $K-1$ intermediate nodes $[v_1, v_2, ..., v_{K-1}] \in \texttt{Paths}(u,w,K)$, as follows
\begin{equation}
    \label{eq:generalization-end}
    \begin{split}
   J_{i,i}^{u,w} & = \frac{1}{\sqrt{d_u d_w}} \sum_{[v_1, v_2, ..., v_{K-1}] \in \texttt{Paths}(u,w,K)}{\prod_{i=1}^{K-1}{\frac{1}{d_{v_i}}}}\\
    \end{split}
\end{equation}
This final term is $O((d_u d_w)^{-1/2})$ in terms of the two neighbors degrees.
This shows that high-degree nodes have less impact on the representations of their neighbors, and that the degree normalization is the main reason.
While the normalization is essential to reduce the bias towards nodes with very high degree, e.g., hubs, it can also make GNNs more vulnerable to attacks from nodes with low degrees.

\para{Node distance impact.}
\label{sec:node-distance}
To explain the impact of node distance on attacks, we start by discussing the relationship between GNNs and network communities.
Then we infer the role of the distance in this context.
Communities are densely connected subgraphs, and are very common in empirical networks, e.g., social networks and co-author networks.
The GNN update rule (Equation~\ref{eq:gcn}) is a special form of Laplacian smoothing~\cite{Li2018Deeper}.
Crucially, GNNs assume that
nodes within the same community tend to share the same label and have similar features~\cite{hussain2020impact,Li2018Deeper}.
This indicates that linking nodes from different communities effectively perturbs the GNN accuracy.

Findings in~\cite{bhattacharyya2014community} show that nodes within similar communities tend to have shorter paths between them.
This supports that linking nearby nodes is likely adding intra-community links, while linking more remote nodes is likely adding inter-community links. 
Therefore, we hypothesize that linking distant nodes would result in more effective attacks.

\subsection{Empirical validation}
\label{sec:numerical-validation}
\begin{table}[]
    \footnotesize
    \centering
    \caption{Dataset statistics.}
\begin{tabular}{lrrrr}
\toprule
       \textbf{Dataset} &    \textbf{Nodes} &      \textbf{Edges} & \textbf{Features} & \textbf{Labels} \\
\midrule
 Citeseer~\cite{sen2008collective} &   2,110 &   3,668 &     3,703 &      6 \\
     Cora~\cite{sen2008collective} &   2,485 &   5,069 &     1,433 &      7 \\
  Cora-ML~\cite{mccallum2000automating} &   2,810 &   7,981 &     2,879 &      7 \\
 Polblogs~\cite{adamic2005political} &   1,222 &  16,714 &     1,490 &      2 \\
   Pubmed~\cite{pubmed} &  19,717 &  44,325 &      500 &      3 \\
\bottomrule
\end{tabular}

    \label{tab:datasets}
\end{table}
Next, we empirically verify the hypotheses from the previous analysis on the datasets, summarized in Table~\ref{tab:datasets}.
We perform perturbation by adding edges to the graph following different strategies.
Then we observe the accuracy of training a GNN model on the perturbed graph.
We choose the well-known non-linear GCN~\cite{gcn} model\footnote{\ex{As the reader might notice, the analysis in Section~\ref{sec:analysis} does not only apply to this particular family of GNNs since feature propagation and normalization are necessary components of GNNs. Our work studies SGC models theoretically and GCN models empirically.}} to empirically show that our theoretical analysis of a linearized GNN model extends to a non-linear one.
In the next experiments, we have a budget of $k = \lfloor r \times m \rfloor$ edges to add to the graph, where $r$ is the perturbation rate which we set to $0.05$.

\para{Node degree.}
The first experiment aims to compare linking low-degree nodes to linking high-degree nodes.
We group the nodes into $10$ equal-sized subsets based on their degrees.
For each pair of subsets, we try adding $k$ adversarial edges between random pairs of nodes in the two subsets and observe the GCN accuracy.
We obtain the results in Figure~\ref{fig:injection-experiment} (top).
These results support our discussion (Section~\ref{sec:analysis}) and show an increase in accuracy, i.e., a decrease in attack effectiveness, when linking pairs of high-degree nodes.
As a result, we assume that attacks are more effective when they \textit{link pairs of low-degree nodes}.

\para{Node distance.}
The second experiment aims to compare linking distant pairs of nodes to linking nearby pairs.
We perform this experiment in $10$ trials, with trial $1$ linking nodes with lowest distances and trial $10$ with highest distances.
In each trial, we observe the GCN accuracy after adding $k$ adversarial edges.
In trial $i\in\{1,..,10\}$, for each adversarial edge (to be added), we randomly pick one node $u$ from the graph and attach one end of that edge to $u$.
Then, we group all the nodes in the graph into $10$ equal-sized subsets based on their distance from $u$.
Finally, we link $u$ to a random node in the $i$-th subset.
Figure~\ref{fig:injection-experiment} (bottom) depicts this comparison and shows the accuracy of each trial.
The figure suggests that \textit{linking distant nodes results in more effective attacks} than linking nearby nodes.


%% file: model.tex
\section{\Structack}
\label{sec:model}
In this section, we introduce our attack strategy \Structack (\textbf{Struc}ture-based at\textbf{tack}), built upon the findings from Section~\ref{sec:hypotheses}.
We outline the attacker's goal, capabilities and knowledge, explain the attack strategy, provide a complexity analysis, and discuss insights on the detection of the attack.

\subsection{Attacker's capabilities and restrictions}
In our setting, the attacker aims to minimize the overall GNN accuracy on node classification.
We limit the knowledge of the attacker to the adjacency matrix, as opposed to existing work~\cite{zugner2018nettack,zugner2019metattack,xu2019topology,ma2019rewatt,dai2018rls2v,ma2020practical}.
The attacker has no access to the features or the label of any node.
They also do not have any information about the attacked GNN model or its parameters. 
We assume that the attacker is able to add edges between any pair of nodes\footnote{\ex{This ability might not directly translate to real-world attacks, but it is necessary to study the extent of different attack approaches, including the baselines that we evaluate as well.}} in the graph, up to a limit $k$, called the \textbf{budget}.
As a result, the attack generates a poisoned adjacency matrix $A'$, where $||A-A'||_0 \leq k$.
According to the taxonomy suggested by~\cite{jin2020survey}, our attack is an untargeted (global) poisoning attack on graph structure.

\subsection{Attack strategy}
\label{sec:attack-strategy}
The findings in Section~\ref{sec:hypotheses} show the impact of low node degrees and long node distances in the graph on adversarial attacks.
Following these findings, an efficient strategy to exploit this impact is to (1) \textit{select} nodes with low degrees, and (2) \textit{link} pairs of nodes with high distances.
Node degree is a measure of node centrality, and distance represents one form of node dissimilarity (e.g., Katz similarity~\cite{newman2018networks} gives higher weights to shorter paths).
We generalize node degree and distance to a diverse set of measures of centrality and similarity.
Therefore, \Structack consists of selecting nodes with the lowest \textit{centrality} and linking these nodes so that the \textit{similarity} between linked nodes is minimized.

For a budget $k$, \Structack chooses $2k$ nodes with the lowest centrality.
We then split these nodes into two sets $U_1$ and $U_2$, both of size $k$, based on their centrality, i.e., $U_1$ has the $k$ nodes with lowest centrality. 
Then \Structack finds the matching between nodes in $U_1$ and $U_2$, which minimizes the sum of similarities between the matched nodes.
To solve this minimization problem, we use the Hungarian algorithm.
Finally, \Structack adds edges between matched nodes. 

For selection and linking steps, we investigate different choices of centrality and similarity measures (Table~\ref{tab:complexity}).
Otherwise, we follow conventional procedures, e.g., splitting lowest-centrality nodes in order, and using the sum of similarities as a criterion for the matching problem.
Please note, investigating other splitting and matching criteria can be interesting, e.g., using interleaving splitting.
However, we leave this for future work as we are more interested in the impact of centrality and similarity choices.

\subsection{Complexity analysis}
After computing the centrality for each node, obtaining the $2k$ lowest-centrality nodes for the splitting step requires $\mathcal{O}(n \log k)$ time.
At the final step of \Structack, finding the optimal node matching is a minimum cost maximum bipartite matching problem.
We solve this problem using the Hungarian algorithm, which has the complexity of $\mathcal{O}(k^3)$ time. 

In Table~\ref{tab:complexity}, we list the centrality and similarity measures we used with their corresponding time and memory complexity.
These measures are well defined in the literature, along with their complexity.
However, to make our paper self-contained, we explain essential details about how we compute similarity and give the resulting time complexity.

\para{Community-based similarity:}
First, we perform community detection using Louvain method~\cite{louvain}, which splits the graph into $C$ disjoint communities.
We then build a community similarity matrix $\mathbf{S} \in \RR^{C \times C}$ encoding the original density of edges, i.e., $\mathbf{S}_{i,j}$ represents the edge density of links between community~$i$ and community~$j$.
Then we set the similarity between two nodes $u$ and $v$ to the similarity of their corresponding communities $\mathbf{S}_{Comm(u), Comm(v)}$, where $Comm(x)$ is the community of node $x$ as per Louvain method.
For the community-based similarity, the time complexity of Louvain community detection is considered to be linear in the number of edges on typical and sparse data~\cite{louvain} $\mathcal{O}(m)$, and the edge density computation step is also of order $\mathcal{O}(m)$, making this similarity calculation of order $\mathcal{O}(m)$ as well.

\para{Distance-based similarity:} We use breadth-first search (BFS) to get single-source shortest paths from each node in $U_1$ to all nodes in $U_2$ (which, in the worst case, means to all nodes in the graph). We choose BFS because we assume that the input graph is unweighted as mentioned in Section~\ref{sec:prelim}. We restrict BFS sources to nodes in $U_1$ since the distance between nodes outside $U_1$ and $U_2$ are not relevant for \Structack.
For the shortest path length computation, and if we do not consider parallelization, the BFS algorithm is repeated $k$ times (once for each node in $U_1$), which gives a time complexity of $O(km)$. Please note that a higher distance indicates a lower similarity.

\para{Katz similarity:}
This notion is a measure of regular equivalence of nodes~\cite{newman2018networks} .
It counts paths of all lengths and weighs them differently, i.e., shorter paths with higher weights. 
We can write Katz similarity matrix as $ \sum_{i=0}^{\infty}{(\alpha A)^i}, $  where $\alpha$ is a constant which needs to be less than the inverse of the largest eigenvalue of $A$.
We approximate the similarity matrix without matrix inversion using \textit{inverse iteration} until the matrix converges after $t$ iterations.
With sparse matrix multiplication, the time complexity turns into $\mathcal{O}(t m)$.
The number of iterations goes down to the desired precision of the similarity.
For a typical choice of $\alpha = 0.85$, we obtain a precision of $10^{-6}$ with $t=100$ iterations\footnote{This argument also applies for computing Pagerank which is also $\mathcal{O}(t m)$.}.

\begin{table}[]
    \centering
    \footnotesize
    \caption{Description of considered centrality/similarity metrics with their time and memory complexity. We list abbreviations for the metric names to use later in results tables.
    $n$ and $m$ represent the number of nodes and edges in the graph respectively, and $k$ is the attack budget. $t$ is the number of iterations for computing Pagerank centrality and Katz similarity.}
    \label{tab:complexity}

    \begin{tabular}{ccc}
        \toprule
         Centrality metric & Time complexity & Memory complexity \\
         \midrule
         Degree (DG) & $\mathcal{O}(m)$ & $\mathcal{O}(m)$ \\
         Eigenvector (EV)~\cite{newman2008mathematics} & $\mathcal{O}(m)$ & $\mathcal{O}(n+m)$ \\
         Pagerank (PR)~\cite{page1999pagerank} &  $ \mathcal{O}(tm)$ & $\mathcal{O}(n+m)$\\
         Betweenness (BT)~\cite{newman2008mathematics,brandes2001faster} & $\mathcal{O}(nm)$ & $\mathcal{O}(n+m)$ \\
         Closeness (CL)~\cite{newman2008mathematics,freeman1978centrality} &$\mathcal{O}(nm)$&$\mathcal{O}(n+m)$ \\ 
         \midrule
         \midrule
         Similarity metric & Time complexity & Memory complexity \\
         \midrule
         Katz (Katz)~\cite{newman2018networks} & $\mathcal{O}(tm)$ & $\mathcal{O}(n^2)$\\
         Community-based (Comm) & $\mathcal{O}(m)$ & $\mathcal{O}(m)$\\
         Distance-based (Dist) & $\mathcal{O}(km)$ & $\mathcal{O}(m)$ \\
         \bottomrule
    \end{tabular}

\end{table}

\subsection{Insights on attack detection}
\label{subsec:attack-unnoticeability-consideration}




\Structack selects nodes with low centrality, which typically have few edges.
Therefore, the attack can cause a significant change in the degree distribution. 
We hence suggest observing the changes in the degree distribution, similar to~\cite{zugner2018nettack}.

Moreover, \Structack links pairs of nodes with low structural similarity, which are likely to have few common neighbors.
The local clustering coefficient of a node is lower with fewer edges shared among its neighbors~\cite{newman2018networks}.
Based on that, we expect that \Structack causes a significant change in the local clustering coefficients of the nodes being linked.
Therefore, we also suggest observing the changes in the local clustering coefficient distribution.

We define two criteria to detect the attack by comparing the original and the perturbed graphs.
First, we test if the node degrees of both graphs stem from the same distribution.
Second, we test if the local clustering coefficient values of both graphs also stem from the same distribution.
If values are assumed to stem from the same distribution in both cases, we consider the attack to be \textit{unnoticeable}.

%% file: experiments.tex
\section{Experimental evaluation}
\label{sec:experiments}

\subsection{Adversarial attack evaluation}
The goal of the experimental evaluation is to test the efficacy of \Structack perturbations on GNNs. 
To this end, we evaluate \Structack against informed baseline attacks as well as the random (uninformed) baseline.
\ex{Notice that our attacks as well as the evaluated baselines apply structural perturbations only and not feature perturbations.}
For a perturbation rate $r$, we allow each attack to perturb the graph by adding (or removing in case of some studied baselines) a budget of $k = \lfloor r \times m \rfloor$ edges.
We evaluate each attack on three different criteria: (i) Effectiveness in terms of GNN misclassification rate, (ii) Efficiency in terms of computation time and memory requirements, and (iii) Unnoticeability in terms of changes of degree and clustering coefficient distributions.
With this evaluation, we aim to demonstrate a performance trade-off of these three aspects. 

\subsection{Experimental setup}
\label{subsec:experimental-setup}
We evaluate $24$ different combinations of (\Structack) derived from combining $6$ different possibilities for node selection (including random selection) with $4$ different possibilities for node linking (including random linking) as listed in Table~\ref{tab:complexity}.
We include random selection and random linking to evaluate whether the effectiveness of certain centrality or similarity choices stem from randomness.
We perform the following evaluations on the $5$  datasets described in Table~\ref{tab:datasets}.

\para{Effectiveness.} 
To evaluate effectiveness (misclassification), we train a GNN model on the perturbed graph and report the classification accuracy on its test set.
Aiming for more robust evaluation (inspired by~\cite{shchur2018pitfalls}), we use $5$ different random splits (10\% train, 10\% validation, and 80\% test) for each dataset.
Our GNN model of choice is the well-known GCN~\cite{gcn} model, which we initialize $5$ times with different random weights for each perturbed input graph.
For the effectiveness evaluation, we set the perturbation rate to $0.05$.

\para{Efficiency.}
Another criterion for evaluating adversarial attacks is their ability to efficiently use available resources in terms of computation time and used memory.
More efficient attacks have a lower runtime and use less memory.
Please note that we ran all experiments on a machine running Linux Ubuntu OS version 16.04 with Intel Xeon E5-2630 Processor with 40 CPUs, 256GB RAM, and a dedicated NVIDIA Tesla P100 16GB GPU. For these efficiency experiments, we also set the perturbation rate to 0.05. 
If an attack did not fit into the GPU memory for a particular dataset, we ran it with CPU settings for that dataset.

\para{Unnoticeability.}
To evaluate attack unnoticeability, we run each attack for different perturbation rates $r \in$ \{0.001, 0.002, 0.003, 0.004, 0.005, 0.0075, 0.01, 0.025,0.05, 0.075, 0.10, 0.15, 0.20\}.
We report results in terms of the critical perturbation rate $r_{critical}$, i.e., largest $r$ for which the attack is still deemed \textit{unnoticeable}.
We consider the attack to be unnoticeable if the changes in the node degree and local clustering coefficient values made by the attack are not significant.
A commonly used approach for comparing two node degree distributions is the Two-Sample Kolmogorov-Smirnov statistical test (KS test)~\cite{aliakbary2014quantification}.
Therefore, we use the KS test to determine whether two compared samples (original graph versus perturbed graph) stem from the same distribution.
We apply this test to obtain the significance in change for both degree and local clustering coefficient distributions.
Here the null hypothesis of the KS test is that \textit{two samples are drawn from the same continuous distribution}.
We set the probability of rejecting the null hypothesis $\alpha$ to $0.05$.


\para{Baselines.}
We evaluate the most effective combinations of \Structack against the following baselines in terms of the three evaluation criteria.

\textbf{Random:} A simple uninformed baseline attack that selects random node pairs and adds an edge between them. This is the only \textit{uninformed} baseline against which we compare \Structack.


\textbf{DICE~\cite{waniek2018hiding}:} A simple heuristic, which is explicitly based on disconnecting nodes with the same label and connecting nodes with different labels. This attack is \textit{informed} as it has access to node labels.

\textbf{Metattack~\cite{zugner2019metattack}:} State-of-the-art optimization-based attack on graphs via meta-learning. It treats the adjacency matrix as a parameter of the optimization problem, which is minimizing the accuracy of a surrogate model. Metattack does not require access to the GNN model parameters, and uses the surrogate model instead.

\textbf{PGD and MinMax~\cite{xu2019topology}:} State-of-the-art optimization-based attacks on graphs. Both attacks apply projected gradient descent to solve the optimization problem after convex relaxation. MinMax attempts to build a more robust attack through attacking a re-trainable GNN. These two attacks require access to the GNN model parameters.


In addition to the graph structure, Metattack, PGD, and MinMax have access to the feature vectors of all nodes and the labels of some nodes (typically, nodes in the training set).
Thus, these three attacks are \textit{informed} in our definition.
These attacks involve randomization, which is why we initialize each of them $5$ times with different random weights for each attack setting.

\begin{figure*}[t]
    \centering
    \includegraphics[width=.9\linewidth]{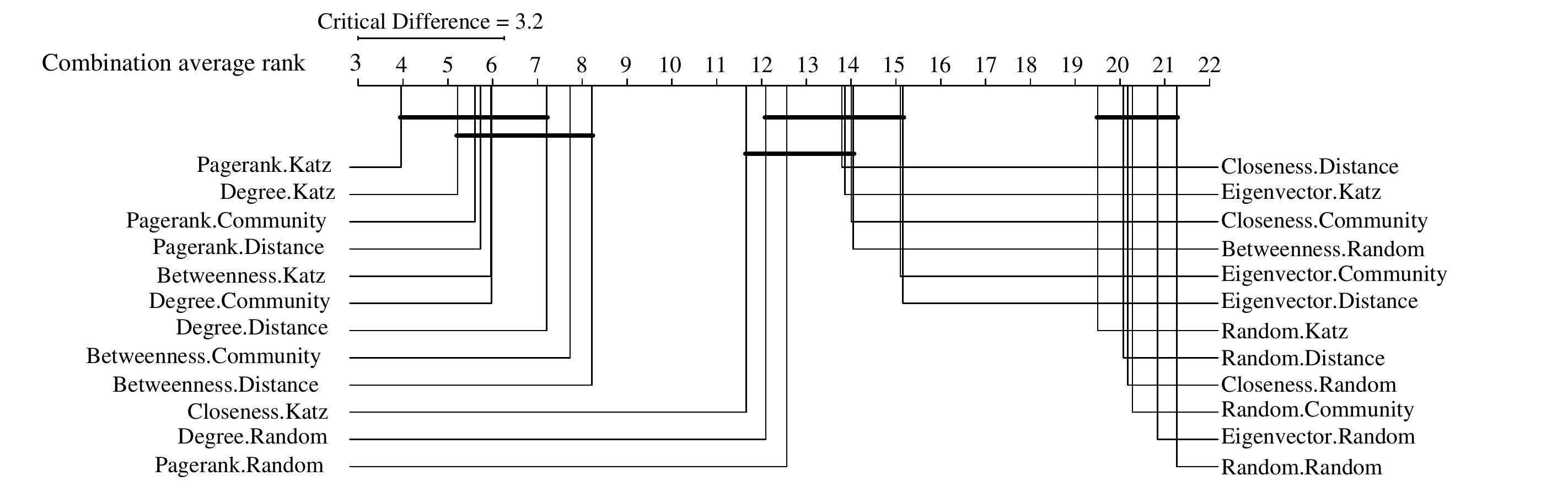} 
    \caption{Comparison of \Structack combinations' effectiveness.
    This plot shows combinations from most to least effective (lowest to highest GCN classification accuracy) presented from left to right. Thick horizontal bars represent no significant difference between the combinations they mark. We find that the best seven combinations are not significantly different, while being significantly better than the rest. We also see that the stronger impact lies in the choice of centrality with the degree and Pagerank centralities with random linking outperforming half of the other combinations. 
    }
    \label{fig:critical-distance}
\end{figure*}

\section{Results and discussion}
Next we present evaluation results, discuss trade-offs, and outline the limitations of our work. 
In the results tables, we use the abbreviations defined in Table~\ref{tab:complexity} to describe the centrality and similarity measures of \Structack combinations.

\para{Effectiveness.}
First, we apply \Structack combinations to each graph dataset and obtain the GCN accuracy.
Then we compute the average rank of each combination in terms of classification accuracy.
We visualize the ranking in Figure~\ref{fig:critical-distance} with a critical difference diagram.
The thick horizontal bars in this figure group together the combinations with no significant difference\footnote{For details on the computation of significance, we refer to the documentation of the R package \texttt{scmamp} \url{https://cran.r-project.org/web/packages/scmamp/scmamp.pdf}.} in ranks between them.
The six lowest-ranked combinations (which involve randomness) perform significantly worse than the rest.
This confirms that the improvement of \Structack does not stem from randomness.
We observe that the seven most effective combinations are not significantly different.
Among these combinations, we frequently see Pagerank centrality, degree centrality and Katz similarity, which implies the effectiveness of these three measures.
Node centrality in \Structack has a substantial impact on effectiveness, relative to the node similarity.
For example, performing selection with degree or Pagerank centrality and linking at random (Degree.Random and Pagerank.Random in Figure~\ref{fig:critical-distance}) seems to perform better than some combinations that do not involve random linking. 

As the seven most effective combinations do not differ significantly from each other, we consequently compare them to the baselines as presented in Table~\ref{tab:acc-baselines}.
\Structack combinations show a comparable performance to state-of-the-art methods, although they have no access to node attributes.

\begin{table}
    \centering
    \footnotesize
    \caption{Adversarial attack effectiveness. This table gives the accuracy of a GCN model trained on the perturbed graph generated by applying each adversarial attack (lower accuracy $\rightarrow$ more effective attack). \Structack (in boldface) is comparable with the state-of-the-art attacks on most datasets with minimum knowledge. According to a Wilcoxon signed-rank test, each Structack approach is significantly more effective than the uninformed Random approach with $p<0.01$ (after Bonferroni correction).
    \textit{*~Metattack could not run for Pubmed with 16GB GPU, and did not finish with CPU settings after 3 weeks running}
    }
    \label{tab:acc-baselines}
    
    
\begin{tabular}{lllllll}
\toprule
           & Dataset &     Citeseer &         Cora &      Cora-ML &     Polblogs &       Pubmed \\
\midrule
           & Clean &  71.90\tiny{$\pm$1.9} &  83.44\tiny{$\pm$1.1} &  85.11\tiny{$\pm$0.7} &  94.66\tiny{$\pm$1.2} &  86.51\tiny{$\pm$0.3} \\
\midrule
\parbox[t]{2mm}{\multirow{4}{*}{\rotatebox[origin=c]{90}{{Informed}}}} & DICE &  70.63\tiny{$\pm$1.8} &  80.09\tiny{$\pm$2.1} &  80.74\tiny{$\pm$2.4} &  82.44\tiny{$\pm$6.1} &  83.32\tiny{$\pm$2.6} \\
           & Metattack &  69.21\tiny{$\pm$1.7} &  76.83\tiny{$\pm$1.3} &  80.01\tiny{$\pm$1.0} &  76.98\tiny{$\pm$0.9} &  N/A* \\
           & MinMax &  68.95\tiny{$\pm$0.8} &  78.45\tiny{$\pm$1.1} &  83.39\tiny{$\pm$0.5} &  85.02\tiny{$\pm$2.2} &  84.97\tiny{$\pm$0.6} \\
           & PGD &  63.80\tiny{$\pm$0.9} &  75.15\tiny{$\pm$1.4} &  80.17\tiny{$\pm$0.7} &  83.28\tiny{$\pm$3.3} &  82.58\tiny{$\pm$0.4} \\
\cline{1-7}
\parbox[t]{2mm}{\multirow{8}{*}{\rotatebox[origin=c]{90}{{Uninformed}}}} & Random &  72.64\tiny{$\pm$1.2} &  80.56\tiny{$\pm$0.5} &  82.24\tiny{$\pm$0.6} &  83.57\tiny{$\pm$3.4} &  85.90\tiny{$\pm$0.3} \\
           & \textbf{BT*Katz} &  71.83\tiny{$\pm$1.0} &  78.77\tiny{$\pm$0.5} &  80.27\tiny{$\pm$0.7} &  76.41\tiny{$\pm$1.8} &  84.38\tiny{$\pm$0.2} \\
           & \textbf{DG*Comm} &  71.89\tiny{$\pm$0.9} &  78.51\tiny{$\pm$0.6} &  80.12\tiny{$\pm$0.6} &  75.65\tiny{$\pm$1.3} &  84.79\tiny{$\pm$0.3} \\
           & \textbf{DG*Dist} &  71.66\tiny{$\pm$1.0} &  78.80\tiny{$\pm$0.5} &  80.15\tiny{$\pm$0.5} &  78.87\tiny{$\pm$2.0} &  84.55\tiny{$\pm$0.3} \\
           & \textbf{DG*Katz} &  71.33\tiny{$\pm$1.1} &  78.98\tiny{$\pm$0.5} &  80.17\tiny{$\pm$0.6} &  76.27\tiny{$\pm$1.4} &  84.34\tiny{$\pm$0.3} \\
           & \textbf{PR*Comm} &  71.67\tiny{$\pm$1.0} &  78.85\tiny{$\pm$0.5} &  80.51\tiny{$\pm$0.6} &  75.25\tiny{$\pm$1.4} &  84.54\tiny{$\pm$0.4} \\
           & \textbf{PR*Dist} &  71.38\tiny{$\pm$1.0} &  78.53\tiny{$\pm$0.5} &  80.19\tiny{$\pm$0.6} &  78.09\tiny{$\pm$2.2} &  84.20\tiny{$\pm$0.3} \\
           & \textbf{PR*Katz} &  71.67\tiny{$\pm$1.0} &  78.40\tiny{$\pm$0.5} &  80.06\tiny{$\pm$0.7} &  75.99\tiny{$\pm$1.6} &  84.08\tiny{$\pm$0.3} \\
\bottomrule
\end{tabular}

\end{table}

\para{Efficiency.}
In Tables~\ref{tab:runtime} and~\ref{tab:memory}, we respectively show the runtime and the memory consumption of our most effective \Structack combinations and existing adversarial attack methods.
We notice a significant drop in runtime and memory consumption for \Structack compared to the optimization-based attacks (Metattack, PGD, and MinMax).
These three attacks did not fit in the available GPU memory for Pubmed, and therefore we ran them with CPU settings for this dataset.
For \Structack combinations, the similarity measure generally has a substantial effect on runtime and memory consumption, with community-based similarity being the most efficient.
An exception to this rule is the runtime of Betweenness and Closeness centralities.
For example on Pubmed, Betweenness and Closeness computation takes $325$ and $66$ minutes respectively, while the computation of Katz similarity takes $100$ minutes.
The time complexity of computing these two measures (Table~\ref{tab:complexity}) is $\mathcal{O}(nm)$ making them impractically slow for large graphs.

\begin{table}
    \centering
    \footnotesize
    \caption{Runtime in minutes with 0.05 perturbation rate. \Structack is in boldface. 
    }
    \label{tab:runtime}

\begin{tabular}{llrrrrr}
\toprule
          & Dataset & Citeseer &   Cora & Cora-ML & Polblogs &   Pubmed \\
\midrule
\parbox[t]{2mm}{\multirow{4}{*}{\rotatebox[origin=c]{90}{{Informed}}}} & DICE &     0.05 &   0.07 &    0.13 &     0.08 &     3.07 \\
          & Metattack &     7.75 &   7.65 &   22.38 &     8.80 &      N/A \\
          & MinMax &    12.83 &  13.03 &   13.68 &    12.58 &  2,645.87 \\
          & PGD &    12.15 &  12.08 &   12.35 &    11.10 &  1,569.55 \\
\cline{1-7}
\parbox[t]{2mm}{\multirow{7}{*}{\rotatebox[origin=c]{90}{{Uninformed}}}} & \textbf{BT*Katz} &     3.33 &   5.12 &    7.42 &     3.87 &   379.00 \\
          & \textbf{DG*Comm} &     0.03 &   0.05 &    0.10 &     0.25 &     1.70 \\
          & \textbf{DG*Dist} &     0.05 &   0.10 &    0.23 &     0.82 &     8.08 \\
          & \textbf{DG*Katz} &     0.98 &   1.30 &    1.55 &     0.63 &    97.98 \\
          & \textbf{PR*Comm} &     0.05 &   0.08 &    0.15 &     0.27 &     1.90 \\
          & \textbf{PR*Dist} &     0.10 &   0.12 &    0.27 &     0.87 &     8.13 \\
          & \textbf{PR*Katz} &     0.93 &   1.32 &    1.52 &     0.72 &    93.28 \\
\bottomrule
\end{tabular}

\end{table}

\begin{table}
    \centering
    \footnotesize
    \caption{Memory consumption in Megabytes with 0.05 perturbation rate. \Structack is in boldface. 
    \textit{*~Snapshot taken after 3 weeks of running}.
    }
    \label{tab:memory}

\begin{tabular}{llrrrrr}
\toprule
          & Dataset & Citeseer & Cora & Cora-ML & Polblogs & Pubmed \\
\midrule
\parbox[t]{2mm}{\multirow{4}{*}{\rotatebox[origin=c]{90}{{Informed}}}} & DICE &      313 & 1,623 &    1,213 &     1,230 &    773 \\
          & Metattack &     2,074 & 2,096 &    2,123 &     2,078 &    *58,394 \\
          & MinMax &     2,176 & 2,243 &    2,318 &     2,109 &  20,554 \\
          & PGD &     2,155 & 2,232 &    2,299 &     2,110 &  19,779 \\
\cline{1-7}
\parbox[t]{2mm}{\multirow{7}{*}{\rotatebox[origin=c]{90}{{Uninformed}}}} & \textbf{BT*Katz} &      578 &  626 &     677 &      442 &  13,918 \\
          & \textbf{DG*Comm} &      316 &  322 &     337 &      403 &    901 \\
          & \textbf{DG*Dist} &      433 &  431 &     430 &      433 &   1,995 \\
          & \textbf{DG*Katz} &      556 &  587 &     662 &      440 &  13,918 \\
          & \textbf{PR*Comm} &      445 &  443 &     443 &      460 &    928 \\
          & \textbf{PR*Dist} &      445 &  443 &     442 &      450 &   2,021 \\
          & \textbf{PR*Katz} &      570 &  617 &     668 &      441 &  13,919 \\
\bottomrule
\end{tabular}

\end{table}

\para{Unnoticeability.} 
We report the critical perturbation rate $r_{critical}$ for which the respective attack remains unnoticeable as per our definition in Sections~\ref{subsec:experimental-setup}.
We present $r_{critical}$ for each approach in Table~\ref{tab:unnoticeability}.
For most datasets, \Structack's $r_{critical}$ is on par or slightly lower than the informed approaches.
We also observe that the choice for node selection strategy in \Structack has a greater influence on the attack unnoticeability than the node linking strategy. 


\begin{table}
    \centering
    \footnotesize
    \caption{Maximum unnoticeable perturbation rate. We evaluate unnoticeability in terms of critical perturbation rate $r_{critical}$, which we define as as the maximum perturbation rate for which the attack remains unnoticeable as per definition in Section~\ref{subsec:experimental-setup}.
    We present the results for $r_{critical}$ per dataset and per adversarial attack. \Structack is in boldface.
    }
\begin{tabular}{lllllll}
\toprule
          & Dataset &     Citeseer &         Cora &      Cora-ML &     Polblogs &       Pubmed \\
\midrule
\parbox[t]{2mm}{\multirow{4}{*}{\rotatebox[origin=c]{90}{\scriptsize{Informed}}}} & DICE & 0.0750 & 0.0500 & 0.0500 & 0.0100 & 0.0100 \\
          & Metattack   & 0.0100 & 0.0100 & 0.0100 & 0.0040 & N/A \\
          & MinMax      & 0.0750 & 0.0750 & 0.0750 & 0.0500 & 0.0040 \\
          & PGD         & 0.1000 & 0.0750 & 0.0500 & 0.1000 & 0.0040 \\
\cline{1-7}
\parbox[t]{2mm}{\multirow{8}{*}{\rotatebox[origin=c]{90}{\scriptsize{Uninformed}}}}  & Random & 0.0250 & 0.0250 & 0.0250 & 0.0100	& 0.0050 \\
          & \textbf{BT*Katz} & 0.0100 & 0.0100 & 0.0075 & 0.0020 & 0.0030 \\
          & \textbf{DG*Comm} & 0.0100 & 0.0075 & 0.0050 & 0.0020 & 0.0030 \\
          & \textbf{DG*Dist} & 0.0100 & 0.0075 & 0.0050 & 0.0020 & 0.0030 \\
          & \textbf{DG*Katz} & 0.0100 & 0.0075 & 0.0050 & 0.0020 & 0.0030 \\
          & \textbf{PR*Comm} & 0.0100 & 0.0075 & 0.0050 & 0.0020 & 0.0030 \\
          & \textbf{PR*Dist} & 0.0100 & 0.0075 & 0.0050 & 0.0020 & 0.0030 \\
          & \textbf{PR*Katz} & 0.0100 & 0.0075 & 0.0050 & 0.0020 & 0.0030 \\
\bottomrule
\end{tabular}
\label{tab:unnoticeability}

\end{table}

%% file: discussion.tex


\para{Performance trade-off.}
\Structack provides competitive effectiveness and high efficiency.
However, it shows to be relatively noticeable compared to baseline approaches.
On the other hand, optimization-based informed attacks achieve better unnoticeability but with much lower efficiency compared to \Structack.
This low efficiency prevents them from running on larger graphs, with Pubmed as a toy example (this has been recently noted by \citet{geisler2021attacking}).
A deeper look into \Structack shows that the selection strategy (i.e., centrality measure) has more impact on effectiveness and unnoticeability.
Conversely, the linking strategy (i.e., similarity measure) has more impact on the efficiency.

All in all, we assume an attack to be effective (cause high misclassification rate) if one of the 7 most effective combinations is picked.
When running on big graphs, attackers would tend to choose efficient combinations such as DG$\times$Comm.
To hide their behavior, attackers would tend to choose less noticeable combinations such as BT$\times$Katz.


\para{Limitations.}
Our study focuses on exploiting the structure information using centrality and similarity measures.
One could study other centrality and similarity measures, and even other graph structural properties. 
Moreover, instead of the theoretical strategy defined in Section~\ref{sec:attack-strategy}, one could define a more practical heuristic to exploit these structural features.
Furthermore, other forms of degree normalization in the target GNN model could result in different strategies than \Structack, which is an interesting direction for future work.
However, the aim of our work is to illustrate the extent to which uninformed attacks are successful, and we demonstrate that through our \Structack strategy, which covers a range of possibilities of uninformed attacks.

Our unnoticeability measure was limited to degree and clustering coefficient distributions. Different unnoticeability tests could be investigated for this purpose.
In this regard, \Structack appears more noticeable than existing informed attacks due to its greediness in selecting nodes with lowest centrality.
The unnoticeability results motivate us to look into approaches that intrinsically consider both effectiveness and unnoticeability.
More careful selection could improve \Structack's unnoticeability, at the possible cost of effectiveness.

Additionally, comparing distributions of the clean graph and the perturbed one is not practical for dynamic networks, where edges and nodes are added and removed constantly.
This comparison does not consider the natural growth of the network.
This type of comparison is a common practice in works on adversarial attacks on graphs, and it should be improved.
This could be alleviated by using graph growth models or dedicated datasets with edge timestamps.


%% file: related-work.tex
\section{Related work}

\para{Information available to attackers.}
Many recent works have introduced attack models for GNNs with different knowledge and capabilities. These models adhere to various restrictions on the practicality of the adversarial attacks and the limitations of the attacker.
However, the majority of these models assume the attacker's knowledge of the targeted GNN model~\cite{xu2019topology,wu2019jaccard} or their access to node attributes~\cite{ma2020practical,zugner2019metattack,zugner2018nettack,sun2019node}, i.e., feature vectors and some labels.
We have referred to such adversarial attacks as \textit{informed attacks}.
A recent survey~\cite{jin2020survey} describes the level of knowledge of (i) the targeted GNN model and (ii) graph data as one characteristic of the attack.
Our work differentiates between these two descriptions and focuses on the knowledge of graph data regardless of the knowledge of the targeted model.

\para{Node centrality.}
Earlier findings in network science on controlling complex networks~\cite{liu2011controllability} show that fewer nodes are needed to control the network, if one aims to control nodes with low degrees.
Another study about the stability of node embedding~\cite{schumacher2020effects} shows that high-centrality nodes have more stable embeddings compared to low-centrality nodes. 
In the context of GNNs, Metattack\cite{zugner2019metattack} shows a slight tendency to connect nodes with low degree.
\citet{zhu2019robust} experimentally consider attacks on nodes with higher than $10$ degrees for noticeability considerations.
\citet{ma2020practical} introduce practical adversarial attacks by targeting nodes with high importance score, e.g., PageRank, node degree, and betweenness.
The authors argue that nodes with too high importance score, e.g., hubs, are hard to control, hence the attack approach avoids such nodes.
Our work conversely builds theoretical grounds and experimental support to show that attacks are more effective if they focus on low degree nodes.

\para{Node similarity.}
A study on the behavior of GNNs~\cite{Li2018Deeper} shows that feature and label smoothness are the reason why GNNs work.
Some works on GNN adversarial attacks~\cite{jin2020survey,jin2020prognn} analyze the poisoned graphs of popular attack models and show a tendency of the attackers to add edges between nodes with different labels and low-similarity features.
~\citet{waniek2018hiding} introduce an attack that is explicitly based on disconnecting nodes with the same label and connecting nodes with different labels (Disconnect Internally, Connect Externally - DICE).
More insights on structure in Metattack~\cite{zugner2019metattack} suggest that attacks tend to link pairs of nodes with higher-than-average shortest path length.
Finally, a preprocessing-based defense mechanism for GNNs~\cite{wu2019jaccard} is based on reducing the weight of edges between nodes with a low Jaccard similarity score of their features.
Our work builds on these findings to investigate more in structural node similarity and build an uninformed structure-based adversarial attack strategy.

%% file: conclusion.tex
\section{Conclusion}
We investigated the effectiveness of uninformed adversarial attacks on GNNs, i.e. attacks that have no access to information about node labels or feature vectors in a graph.
With theoretical considerations and experimental support, we demonstrated that uninformed attacks can exploit structural features of the graph, such as node centrality and similarity.
We presented \Structack, a novel uninformed attack strategy that selects nodes with low centrality and links pairs of nodes with low similarity.
In experiments on five graph datasets \Structack showed comparable performance to state-of-the-art attacks, while having less information about the graph (no access to node attributes), exhibiting higher efficiency, and reasonable unnoticeability.
Our work shows that uninformed adversarial attacks are successful with only structural knowledge, sometimes outperforming informed attacks.
The feasibility of \Structack on real-world graphs makes it vital to develop more structure-aware defense mechanisms for more reliable GNN prediction.

%% file: acknowledgements.tex
\section*{Acknowledgements}
The Know‐Center is funded within the Austrian COMET Program – Competence Centers for Excellent Technologies – under the auspices of the Austrian Federal Ministry of Transport, Innovation and Technology, the Austrian Federal Ministry of Economy, Family and Youth and by the State of Styria. COMET is managed by the Austrian Research Promotion Agency FFG.
This work is supported by the H2020 project TRUSTS (GA: 871481) and the “DDAI” COMET Module within the COMET Program, funded by the Austrian Federal Ministry for Transport, Innovation and Technology (bmvit), the Austrian Federal Ministry for Digital and Economic Affairs (bmdw), the Austrian Research Promotion Agency (FFG), the province of Styria (SFG) and partners from industry and academia.

%% file: appendix.tex
\section{Detailed effectiveness results}
\label{app:detail-result}
Table shows the accuracy of GCN~\cite{gcn} for node classification on the considered datasets after changing the structure using different combinations of \Structack with a perturbation rate $r=0.05$. These are the detailed results of what Figure~\ref{fig:critical-distance} summarizes.

\begin{table}[h]
\centering
\footnotesize
\caption{GCN accuracy on each dataset after applying \Structack with each centrality$\times$similarity combination. The lowest accuracy (best combination) in each dataset is shown in boldface.} 
\label{tab:acc}
\begin{tabular}{llllll}
\toprule
       & Similarity &    Community &     Distance &         Katz &       Random \\
Dataset & Centrality &              &              &              &              \\
\midrule
\multirow{6}{*}{Citeseer} & Betweenness &  72.06\tiny{$\pm$0.9} &  72.04\tiny{$\pm$1.0} &  71.83\tiny{$\pm$1.0} &  72.22\tiny{$\pm$1.0} \\
       & Closeness &  72.75\tiny{$\pm$0.7} &  71.99\tiny{$\pm$1.0} &  72.22\tiny{$\pm$1.1} &  72.88\tiny{$\pm$1.1} \\
       & Degree &  71.89\tiny{$\pm$0.9} &  71.66\tiny{$\pm$1.0} &  \textbf{71.33\tiny{$\pm$1.1}} &  71.86\tiny{$\pm$0.9} \\
       & Eigenvector &  72.44\tiny{$\pm$1.1} &  72.55\tiny{$\pm$0.8} &  72.46\tiny{$\pm$1.3} &  73.05\tiny{$\pm$0.8} \\
       & Pagerank &  71.67\tiny{$\pm$1.0} &  71.38\tiny{$\pm$1.0} &  71.67\tiny{$\pm$1.0} &  72.18\tiny{$\pm$0.7} \\
       & Random &  73.08\tiny{$\pm$1.1} &  73.08\tiny{$\pm$1.1} &  73.04\tiny{$\pm$0.9} &  72.64\tiny{$\pm$1.2} \\
\cline{1-6}
\multirow{6}{*}{Cora} & Betweenness &  79.05\tiny{$\pm$0.5} &  79.11\tiny{$\pm$0.4} &  78.77\tiny{$\pm$0.5} &  79.65\tiny{$\pm$0.5} \\
       & Closeness &  79.99\tiny{$\pm$0.4} &  79.99\tiny{$\pm$0.6} &  79.57\tiny{$\pm$0.5} &  80.33\tiny{$\pm$0.4} \\
       & Degree &  78.51\tiny{$\pm$0.6} &  78.80\tiny{$\pm$0.5} &  78.98\tiny{$\pm$0.5} &  79.63\tiny{$\pm$0.5} \\
       & Eigenvector &  80.19\tiny{$\pm$0.5} &  79.80\tiny{$\pm$0.5} &  79.93\tiny{$\pm$0.6} &  80.53\tiny{$\pm$0.5} \\
       & Pagerank &  78.85\tiny{$\pm$0.5} &  78.53\tiny{$\pm$0.5} &  \textbf{78.40\tiny{$\pm$0.5}} &  78.99\tiny{$\pm$0.5} \\
       & Random &  80.35\tiny{$\pm$0.5} &  80.44\tiny{$\pm$0.5} &  80.31\tiny{$\pm$0.5} &  80.56\tiny{$\pm$0.5} \\
\cline{1-6}
\multirow{6}{*}{Cora-ML} & Betweenness &  80.50\tiny{$\pm$0.7} &  80.16\tiny{$\pm$0.5} &  80.27\tiny{$\pm$0.7} &  80.85\tiny{$\pm$0.7} \\
       & Closeness &  81.42\tiny{$\pm$0.7} &  81.32\tiny{$\pm$0.7} &  81.58\tiny{$\pm$0.6} &  82.14\tiny{$\pm$0.6} \\
       & Degree &  80.12\tiny{$\pm$0.6} &  80.15\tiny{$\pm$0.5} &  80.17\tiny{$\pm$0.6} &  80.51\tiny{$\pm$0.7} \\
       & Eigenvector &  81.72\tiny{$\pm$0.8} &  81.60\tiny{$\pm$0.5} &  81.48\tiny{$\pm$0.7} &  82.09\tiny{$\pm$0.7} \\
       & Pagerank &  80.51\tiny{$\pm$0.6} &  80.19\tiny{$\pm$0.6} &  \textbf{80.06\tiny{$\pm$0.7}} &  80.99\tiny{$\pm$0.8} \\
       & Random &  82.23\tiny{$\pm$0.7} &  82.00\tiny{$\pm$0.6} &  82.10\tiny{$\pm$0.6} &  82.24\tiny{$\pm$0.6} \\
\cline{1-6}
\multirow{6}{*}{Polblogs} & Betweenness &  \textbf{75.19\tiny{$\pm$0.9}} &  77.88\tiny{$\pm$3.0} &  76.41\tiny{$\pm$1.8} &  83.17\tiny{$\pm$2.5} \\
       & Closeness &  75.87\tiny{$\pm$1.4} &  79.09\tiny{$\pm$2.4} &  75.72\tiny{$\pm$1.8} &  83.02\tiny{$\pm$2.0} \\
       & Degree &  75.65\tiny{$\pm$1.3} &  78.87\tiny{$\pm$2.0} &  76.27\tiny{$\pm$1.4} &  82.48\tiny{$\pm$2.9} \\
       & Eigenvector &  76.52\tiny{$\pm$0.9} &  77.69\tiny{$\pm$2.1} &  76.62\tiny{$\pm$1.7} &  82.41\tiny{$\pm$2.7} \\
       & Pagerank &  75.25\tiny{$\pm$1.4} &  78.09\tiny{$\pm$2.2} &  75.99\tiny{$\pm$1.6} &  82.73\tiny{$\pm$2.6} \\
       & Random &  79.93\tiny{$\pm$3.3} &  80.25\tiny{$\pm$2.9} &  80.13\tiny{$\pm$3.4} &  83.57\tiny{$\pm$3.4} \\
\cline{1-6}
\multirow{6}{*}{Pubmed} & Betweenness &  84.93\tiny{$\pm$0.3} &  84.71\tiny{$\pm$0.2} &  84.38\tiny{$\pm$0.2} &  85.21\tiny{$\pm$0.3} \\
       & Closeness &  85.42\tiny{$\pm$0.3} &  85.38\tiny{$\pm$0.2} &  85.24\tiny{$\pm$0.3} &  85.58\tiny{$\pm$0.3} \\
       & Degree &  84.79\tiny{$\pm$0.3} &  84.55\tiny{$\pm$0.3} &  84.34\tiny{$\pm$0.3} &  85.08\tiny{$\pm$0.4} \\
       & Eigenvector &  85.45\tiny{$\pm$0.3} &  85.49\tiny{$\pm$0.2} &  85.40\tiny{$\pm$0.3} &  85.65\tiny{$\pm$0.2} \\
       & Pagerank &  84.54\tiny{$\pm$0.4} &  84.20\tiny{$\pm$0.3} & \textbf{ 84.08\tiny{$\pm$0.3}} &  85.13\tiny{$\pm$0.2} \\
       & Random &  85.83\tiny{$\pm$0.3} &  85.74\tiny{$\pm$0.3} &  85.64\tiny{$\pm$0.3} &  85.90\tiny{$\pm$0.3} \\
\bottomrule
\end{tabular}

\end{table}

\newpage
\section{Detailed efficiency results}
\label{app:detail-efficiency}
We show the detailed runtime for \Structack combinations in Table~\ref{tab:runtime-all} and the memory consumption in Table~\ref{tab:memory-all}. Runtime and memory consumption results are obtained after setting random linking with each selection method, and random selection with each linking method. We chose random because its time and memory requirements are negligible for our comparison. For these experiments, we also set the perturbation rate $r$ to 0.05.

\begin{table}[h]
    \centering
    \footnotesize
    \caption{Runtime in seconds for each selection/linking method.}
    \label{tab:runtime-all}

\begin{tabular}{lllllll}
\toprule
           & Dataset & Citeseer &    Cora & Cora-ML & Polblogs &    Pubmed \\
\midrule
\multirow{5}{*}{Centrality} & Betweenness &   170.42 &  242.02 &  398.90 &   233.17 &  19,507.80 \\
           & Closeness &    36.95 &   54.19 &   91.24 &    66.79 &   3,938.98 \\
           & \textbf{Degree} &     \textbf{0.18} &    \textbf{0.42} &    \textbf{0.46} &     \textbf{0.96} &      \textbf{2.82} \\
           & Eigenvector &     1.04 &    2.10 &    6.12 &     5.57 &     21.56 \\
           & Pagerank &     2.01 &    2.22 &    3.09 &     4.65 &     20.86 \\
\cline{1-7}
\multirow{3}{*}{Similarity} & \textbf{Community} &     \textbf{2.38} &   \textbf{ 3.06} &   \textbf{ 5.60} &    \textbf{13.48} &    \textbf{110.42} \\
           & Distance &     3.27 &    5.86 &   13.48 &    49.53 &    468.35 \\
           & Katz &    58.75 &   79.47 &   85.60 &    38.30 &   5,978.16 \\
\bottomrule
\end{tabular}

\end{table}

\begin{table}[h]
    \centering
    \footnotesize
    \caption{Memory consumption in Megabytes for each selection/linking method.}
    \label{tab:memory-all}

\begin{tabular}{llrrrrr}
\toprule
           & Dataset & Citeseer & Cora & Cora-ML & Polblogs & Pubmed \\
\midrule
\multirow{5}{*}{Centrality} & Betweenness &      367 &  366 &     363 &      364 &    438 \\
           & Closeness &      369 &  367 &     365 &      365 &    438 \\
           & \textbf{Degree} &      \textbf{316} &  \textbf{318} &     \textbf{321} &      \textbf{323} &    \textbf{423} \\
           & Eigenvector &      357 &  354 &     353 &      361 &    437 \\
           & Pagerank &      349 &  349 &     354 &      367 &    500 \\
\cline{1-7}
\multirow{3}{*}{Similarity} & \textbf{Community} &      \textbf{368} &  \textbf{367} &     \textbf{370} &      \textbf{387} &   \textbf{1,055} \\
           & Distance &      387 &  382 &     400 &      409 &   2,150 \\
           & Katz &      549 &  609 &     659 &      432 &  1,3881 \\
\bottomrule
\end{tabular}

\end{table}
\newpage
\section{Detailed unnoticeability results}
\label{app:detail-unnoticeability}

We show the unnoticeability of different combinations of \Structack and with different perturbation rates $r \in$ \{0.001, 0.002, 0.003, 0.004, 0.005, 0.0075, 0.01, 0.025,0.05, 0.075, 0.10, 0.15, 0.20\}. These are the detailed results of what Figure~\ref{tab:unnoticeability} summarizes.

Notice that the choice of similarity has no impact on $r_{critical}$.
For Citeseer, Cora and Cora-ML, closeness centrality and eigenvector centrality are the least noticeable.
For the other two datasets (Polblogs and Pubmed), all centrality measures have the same $r_{critical}$

\begin{table}[h]
\centering
\footnotesize
\caption{Attack unnoticeability on each dataset after applying \Structack with each centrality$\times$similarity combination. 
 The centrality measure (excpet random) with the highest critical perturbation rate $r_{critical}$ in each dataset is shown in boldface.} 
\label{tab:detailed-unnoticeability}
\begin{tabular}{llllll}
\toprule
       & Similarity &    Community &     Distance &         Katz &       Random \\
Dataset & Centrality &              &              &              &              \\
\midrule
\multirow{6}{*}{Citeseer} & Betweenness & 0.0100 & 0.0100 & 0.0100 & 0.0100 \\
       & \textbf{Closeness} & 0.0250 & 0.0250 & 0.0250 & 0.0250 \\
       & Degree & 0.0100 & 0.0100 & 0.0100 & 0.0100 \\
       & \textbf{Eigenvector} & 0.0250 & 0.0250 & 0.0250 & 0.0250 \\
       & Pagerank & 0.0100 & 0.0100 & 0.0100 & 0.0100 \\
       & Random & 0.0500 & 0.0500 & 0.0500 & 0.0500 \\
\cline{1-6}
\multirow{6}{*}{Cora} & Betweenness & 0.0100 & 0.0100 & 0.0100 & 0.0100 \\
       & \textbf{Closeness} & 0.0250	& 0.0250 & 0.0250 & 0.0250 \\
       & Degree & 0.0075 & 0.0075 & 0.0075 & 0.0075 \\
       & \textbf{Eigenvector} & 0.0250 & 0.0250 & 0.0250 & 0.0250 \\
       & Pagerank & 0.0075 & 0.0075 & 0.0075 & 0.0075 \\
       & Random & 0.0250 & 0.0250 & 0.0250 & 0.0250 \\
\cline{1-6}
\multirow{6}{*}{Cora-ML} & Betweenness & 0.0100	& 0.0100 & 0.0100 & 0.0100 \\
       & \textbf{Closeness} & 0.0100 & 0.0100 & 0.0100 & 0.0100 \\
       & Degree & 0.0050 & 0.0050 & 0.0050 & 0.0050 \\
       & \textbf{Eigenvector} & 0.0100 & 0.0100 & 0.0100 & 0.0100 \\
       & Pagerank & 0.0050 & 0.0050 & 0.0005 & 0.0050 \\
       & Random & 0.0250 & 0.0250 & 0.0250 & 0.0250 \\
\cline{1-6}
\multirow{6}{*}{Polblogs} & \textbf{Betweenness} & 0.0020 & 0.0020 & 0.0020 & 0.0020 \\
       & \textbf{Closeness} & 0.0020 & 0.0020 & 0.0020 & 0.0020 \\
       & \textbf{Degree} & 0.0020 & 0.0020 & 0.0020 & 0.0020 \\
       & \textbf{Eigenvector} & 0.0020 & 0.0020 & 0.0020 & 0.0020 \\
       & \textbf{Pagerank} & 0.0020 & 0.0020 & 0.0020 & 0.0020 \\
       & Random & 0.0100 & 0.0100 & 0.0100 & 0.0100 \\
\cline{1-6}
\multirow{6}{*}{Pubmed} & \textbf{Betweenness} & 0.0030 & 0.0030 & 0.0030 & 0.0030 \\
       & \textbf{Closeness} & 0.0030 & 0.0030 & 0.0030 & 0.0030 \\
       & \textbf{Degree} & 0.0030 & 0.0030 & 0.0030 & 0.0030 \\
       & \textbf{Eigenvector} & 0.0030 & 0.0030 & 0.0030 & 0.0030 \\
       & \textbf{Pagerank} & 0.0030 & 0.0030 & 0.0030 & 0.0030 \\
       & Random & 0.0050 & 0.0050 & 0.0050 & 0.0050 \\
\bottomrule
\end{tabular}

\end{table}